\pdfoutput=1
\documentclass[11pt]{article}

\usepackage{amsmath,graphicx,makecell,url}
\usepackage{acl}
\usepackage{multirow}% Standard package includes
\usepackage{times}
\usepackage{latexsym}

\usepackage{times}
\usepackage{color}
\usepackage{latexsym}
\usepackage{amsmath}
\usepackage{booktabs}
\usepackage{multirow}
\usepackage{graphics}
\usepackage{graphicx} 
\usepackage{caption}
\usepackage{amssymb}
\usepackage{pifont}
\usepackage{subfigure}
\usepackage{makecell}
\usepackage{float}
\usepackage{stfloats}

\usepackage[T1]{fontenc}
\usepackage[utf8]{inputenc}

\usepackage{microtype}

\title{ConaCLIP: Exploring Distillation of Fully-Connected Knowledge Interaction Graph for Lightweight Text-Image Retrieval}

\author{Jiapeng Wang$^1$\thanks{\ \ Contribution during internship at Alibaba Group.}  \quad  Chengyu Wang$^2$\thanks{\ \ Co-corresponding authors.}  \quad    Xiaodan Wang$^3$   \quad Jun Huang$^2$ \quad  Lianwen Jin$^{1}$\footnotemark[2]\\ 
$^1$South China University of Technology, Guangzhou, China\\
$^2$Alibaba Group, Hangzhou, China\\
$^3$Fudan University, Shanghai, China\\
$^{1}$\texttt{eejpwang@mail.scut.edu.cn,  eelwjin@scut.edu.cn} \\
$^{2}$\texttt{\{chengyu.wcy, huangjun.hj\}@alibaba-inc.com} \\  
$^{3}$\texttt{xiaodanwang20@fudan.edu.cn}\\
}

\begin{document}
\maketitle
\begin{abstract}
Large-scale pre-trained text-image models with dual-encoder architectures (such as CLIP \cite{clip}) are typically adopted for various vision-language applications, including text-image retrieval.
However, these models are still less practical on edge devices or for real-time situations, due to the substantial indexing and inference time and the large consumption of computational resources.
Although knowledge distillation  techniques have been widely utilized for uni-modal model compression, 
how to expand them to the situation when the numbers of modalities and teachers/students are doubled has been rarely studied. 
In this paper, we conduct comprehensive experiments on this topic and propose the fully-\textbf{C}onnected  kn\textbf{o}wledge  interactio\textbf{n}  gr\textbf{a}ph (Cona) technique
for cross-modal pre-training distillation.
Based on our findings,  the resulting ConaCLIP achieves SOTA performances on  the widely-used Flickr30K and MSCOCO benchmarks under the lightweight setting.
An industry application of our method on an e-commercial platform further demonstrates the significant effectiveness 
 of ConaCLIP.\footnote{Related resources will be publicly available in the EasyNLP framework~\cite{DBLP:conf/emnlp/WangQZLLWWHL22}. URL:~\url{https://github.com/alibaba/EasyNLP}.}
\end{abstract}

\section{Introduction}
Text-image retrieval (TIR) aims at  retrieving a list of the most relevant images from a large 
 image collection when a specific text query is given.
 With the rapid development of information interaction and social intercourse, 
 it has been regarded as a crucial component of cross-modal applications and required by various real-world scenarios, such as e-commercial platforms (sites).    
 
Recently, inspired by the great success of pre-trained language models 
\cite{devlin2019bert,liu2019roberta,gpt3},  research on large-scale vision-language pre-training  \cite{tan2019lxmert,li2020oscar,clip,blip,wang2022ofa,DBLP:conf/wsdm/WangLLWZWHX23}  has achieved remarkable progress on a variety of vision-language tasks, including text-image retrieval. These existing methods 
can be typically classified into two categories according to the model architecture: \textit{cross-encoder} and \textit{dual-encoder}. 
\textit{Cross-encoder} typically adds additional Transformer \cite{transformer} layers to model the deep interaction between image and text representations.  It can generally boost the retrieval performance, while resulting in an unbearably slow retrieval speed when applied to the entire image collection since the cross-modal costs are required for each image sample whenever a new text query is given.
In contrast, \textit{dual-encoder} encodes the visual and textual inputs in a wholly decoupled manner.
The image representation is allowed to be pre-computed and re-used independent of the text queries.
Such approaches can also utilize fast approximate nearest neighbor (ANN) search \cite{muja2009fast,jegou2010product,johnson2019billion} at runtime.

\begin{figure}[t]
\centering
\includegraphics[width=0.5\textwidth]{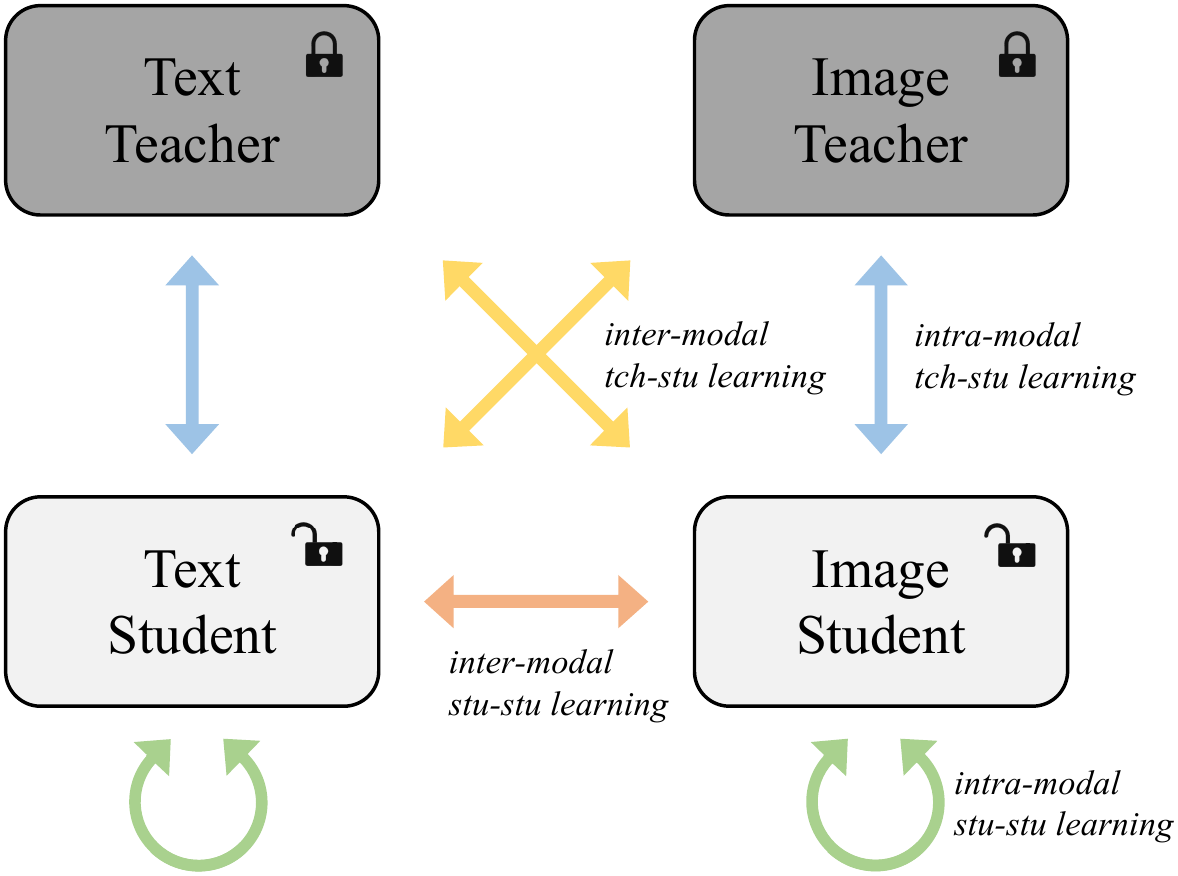}
\caption{Our dual-encoder pre-training distillation framework with Cona. 
Each color of two-way arrows represents a type of knowledge interaction learning. At this stage, the teacher encoders are frozen.
} 
\label{pipeline}
\end{figure}

Although dual-encoder is  usually preferred for real-world applications, the existing related models such as CLIP \cite{clip} are still less practical on edge devices with limited computing resources, or for the dynamic indexing scenario, e.g., private photos/messages collections (sites). 
To address this issue, we aim to start from the large-scale pre-trained dual-encoder models 
and focus on the pre-training distillation to present a series of much smaller, faster, and effective counterparts. 
Knowledge distillation (KD) \cite{Hintonkd} is proposed to transfer knowledge with soft targets from  a teacher to a student  in the same modality.  MoTIS \cite{motis} simply  repeats intra-modal InfoNCE-based \cite{infonce} learning in both  text and image domains
for distillation.
Nevertheless, 
when the number of modalities  
doubles for dual-encoder structure, which means text and image teachers as well as text and image students, these methods still only involve  intra-modal teacher-student knowledge interaction learning. 
Instead, in this paper, we  comprehensively explore the fully-\textbf{C}onnected  kn\textbf{o}wledge  interactio\textbf{n}  gr\textbf{a}ph (Cona) between every possible teacher-student or student-student pair. 
As shown in Fig.~\ref{pipeline}, each two-way arrow represents the knowledge interaction learning between the two models it points to. 
And the aforementioned KD and MoTIS belong to a single \textit{blue} arrow and the two \textit{blue} arrows, respectively.
Moreover, 
in order to better explore the potential of Cona,
we implement and investigate various supervision strategies to guide the model optimization, 
which finally makes each type of learning contribute to the overall improvement.

We release various sizes of lightweight dual-encoder models named ConaCLIP
for 
different real-world scenarios. 
 Compared with the previous SOTA method \cite{motis}, our ConaCLIP achieves 10.6/12.9/12.8 R@1 gains on Flickr30K/MSCOCO (1K)/MSCOCO (5K) benchmarks under the same model setting.
We have also verified its effectiveness  on an  e-commerce platform. 
It can achieve 1.44$\times$/1.92$\sim$4.86$\times$ inference speed-up with competitive performances given image/text queries.
The main  contributions of this paper can be summarized as follows:
\begin{itemize}
\item We propose a new pre-training distillation method with the fully-connected knowledge interaction graph (Cona) for lightweight
dual-encoder models.

\item We release a series of lightweight ConaCLIP models to the open-source community, which can significantly surpass previous SOTA models on the widely-used Flickr30K and  MSCOCO   benchmarks.

\item We provide a real-world application of this method in real industrial scenarios to further demonstrate its practical values.

\end{itemize}

\section{Related Work}
Cross-encoder \cite{tan2019lxmert,li2019visualbert,chen2020uniter,li2020oscar,chen2022improving} refers to multiple layers of dense cross-modal interactions, e.g., cross-attention \cite{transformer}, are typically employed to 
image and text representations for more fine-grained merge and alignment.
Although it often achieves superior retrieval accuracy thanks to the patch/token-level integration, the high memory cost and  computation inefficiency make it impractical under time-critical real-world settings. 

Oppositely, for dual-encoder \cite{zhang2020devlbert,align,clip,dou2022empirical}, image and text features are encoded into a joint embedding  space separately, and the modality interaction is only handled by a simple cosine similarity of the final image and text feature vectors. 
Such approaches can be regarded as scalable and indexable: the specific choices
of encoder architectures can be independent and dynamic, and the late-interaction scheme  allows for efficient large-scale searching. 

Pre-training distillation 
for lightweight dual-encoder architecture
has been rarely studied. Vanilla knowledge distillation \cite{Hintonkd} can be referred to as the knowledge  transfer  from  a teacher to a student  in the same modality based on soft targets. However, it is a general procedure without awareness and pertinence for cross-modal learning. 
MoTIS \cite{motis}  separately compresses text or image encoder with an intra-modal contrastive objective that aligns the output embeddings of the student and teacher of each modality, which can be seen as an alternative form of knowledge distillation. 
Nevertheless, these methods ignore or do not find an appropriate approach to leverage the cross-modal distillation process.
Further than them, our method is dedicated to  exploring the fully-connected knowledge interaction graph for  dual-encoder distillation, which is a natural and effective extension.

\section{Methodology}
In this section, we first give the preliminary knowledge,
then propose our pre-training distillation framework with Cona. 
Finally, we introduce various supervision strategies.
% model learning.

\subsection{Preliminary}
For the sake of explanation, we abbreviate \textit{text}, \textit{image}, \textit{teacher} and \textit{student} as $T$, $I$, $\mathrm{tch}$ and $\mathrm{stu}$ respectively. 
$F$ represents the L2-normalized   feature vector outputted by   the encoder architecture $E$.

Before student learning, the teachers $E^{T}_{\mathrm{tch}}$ and $E^{I}_{\mathrm{tch}}$ are commonly first pre-trained using an objective that pushes the embeddings of matched text-image pairs closer while pushing those of non-matched ones apart, with large model capacity and massive data. Specifically, CLIP \cite{clip} takes the InfoNCE \cite{infonce} loss as the supervision form. Without losing generality, given two outputted feature vectors  $F^a$ and $F^b\in\mathcal{R}^{N\times d}$, we define that:
\begin{align}
p_{i,j}(F^a,F^b)&=\frac{exp(F^a_i {F^b_j}^{\top}/\tau)}{\sum_{k}exp(F^a_i {F^b_k}^{\top}/\tau)},\label{equal_p}\\
\mathcal{L}^{\mathrm{InfoNCE}}_{F^a\rightarrow F^b}& = -\frac{1}{N}\sum_{i=1}^{N}{log(p_{i,i}(F^a,F^b))},\label{equal_infonce}
\end{align}
where $N$ is the mini-batch size, $d$ is the channel size and $\tau$ is the temperature hyper-parameter. The final loss of CLIP can be formulated as:
\begin{align}
    \mathcal{L}^{\mathrm{CLIP}} = \mathcal{L}^{\mathrm{InfoNCE}}_{F^{T}_{\mathrm{tch}}\rightarrow F^{I}_{\mathrm{tch}}} + \mathcal{L}^{\mathrm{InfoNCE}}_{F^{I}_{\mathrm{tch}}\rightarrow F^{T}_{\mathrm{tch}}}.
\end{align}\label{clip}

Next, the pre-training distillation of students $E^{T}_{\mathrm{stu}}$ and $E^{I}_{\mathrm{stu}}$ begins, with parameters of teachers $E^{T}_{\mathrm{tch}}$ and $E^{I}_{\mathrm{tch}}$ frozen. 
% The previous SOTA 
MoTIS \cite{motis} also adopts the InfoNCE-based loss at this stage, and implements it 
in both text and image domains separately:
\begin{align}
    \mathcal{L}^{\mathrm{MoTIS}} = \mathcal{L}^{\mathrm{InfoNCE}}_{F^{T}_{\mathrm{stu}}\rightarrow F^{T}_{\mathrm{tch}}} + \mathcal{L}^{\mathrm{InfoNCE}}_{F^{I}_{\mathrm{stu}}\rightarrow F^{I}_{\mathrm{tch}}}.\label{equal_motis}
\end{align}
According to  the subscript in Eq.~(\ref{equal_motis}), 
it is easy to see that MoTIS only involves
\textit{intra-modal} \textit{teacher-student} learning.

\subsection{Pre-training Distillation with Cona}
Unlike existing works, our method introduces the fully-connected knowledge interaction graph (Cona) for pre-training distillation.
 Apart from  \textit{intra-modal teacher-student} learning, our method also includes \textit{intra-modal student-student} learning, \textit{inter-modal teacher-student} learning and \textit{inter-modal student-student} learning, as shown in Fig.~\ref{pipeline}.
This fully-connected learning graph  established for students  $E^{T}_{\mathrm{stu}}$ and $E^{I}_{\mathrm{stu}}$ serves as an integration of multi-view and multi-task learning schemes, which can strengthen the robustness and effectiveness \cite{caruana1997multitask,luong2015multi,aghajanyan2021muppet} required by pre-trained models.

We suggest that each type of learning process in Cona should be concretely implemented in detailed supervision strategies.
Therefore, we propose and investigate various supervision strategies in the next subsection. 
  
\subsection{Supervision Strategies}
Here we continue to use  $F^a$ and $F^b$ (prediction) along with $\widetilde{F^a}$ and $\widetilde{F^b}$ (target) as placeholders for illustration, and present the following  effective supervision strategies:

\noindent\textbf{InfoNCE loss} is a type of contrastive loss function. It has been formulated in Eq.~(\ref{equal_infonce}), and successfully applied for  distillation by Eq.~(\ref{equal_motis}). 

\noindent\textbf{Feature-wise distance (FD) loss} directly minimizes the distance between feature vectors. We utilize squared L2-norm as the measure:
\begin{align}
\mathcal{L}^{\mathrm{FD}}_{F^a\Leftrightarrow F^b} = \frac{1}{2}\frac{1}{Nd}\sum_{i=1}^{N}{\sum_{j=1}^{d}{(F^a_{i,j}-F^b_{i,j})^2}}.
\end{align}

\noindent\textbf{Similarity-wise distance (SD) loss} minimizes the distance criterion between similarity matrices:
\begin{align}
\resizebox{\hsize}{!}{$\mathcal{L}^{\mathrm{SD}}_{F^a\rightarrow F^b\Leftrightarrow	 \widetilde{F^a}\rightarrow \widetilde{F^b}} = \frac{1}{2}\frac{1}{N^2}\sum\limits_{i=1}\limits^{N}{\sum\limits_{j=1}\limits^{N}{(F^a_i {F^b_j}^{\top}-\widetilde{F^a_i}\widetilde{F^b_j}^{\top})^2}}.$}\label{sd}
\end{align}
Since  $F^a$, $F^b$, $\widetilde{F^a}$ and $\widetilde{F^b}$ have been L2-normalized, the values of cosine-similarities $F^a_i {F^b_j}^{\top}$ and $\widetilde{F^a_i}\widetilde{F^b_j}^{\top}$ are in the range $[-1, 1]$. The distance between prediction $F^a_i {F^b_j}^{\top}$ and target $\widetilde{F^a_i}\widetilde{F^b_j}^{\top}$ needs to be shortened. Hence, the squared L2-norm is also adopted here.

\noindent\textbf{KL-Div loss} uses the Kullback–Leibler divergence to measure the difference between the predicted and the target probability distributions. Given $p_{i,j}$ acquired by softmax operation shown in Eq.~(\ref{equal_p}), it minimizes the following optimization objective:
\begin{align}
\resizebox{\hsize}{!}{$\mathcal{L}^{\mathrm{KL\text{-}Div}}_{F^a\rightarrow F^b\lVert \widetilde{F^a}\rightarrow \widetilde{F^b}} = \frac{1}{N}\sum\limits_{i=1}\limits^{N}{\sum\limits_{j=1}\limits^{N}{p_{i,j}(F^a,F^b)log\frac{p_{i,j}(F^a,F^b)}{p_{i,j}(\widetilde{F^a},\widetilde{F^b})}}}.$}\label{kldiv}
\end{align}

\begin{table*}[!]
\centering
\scalebox{0.64}{
\begin{tabular}{@{}l|ccccccccccc@{}}
\toprule
 \multicolumn{1}{c|}{\multirow{2}{*}{Learning Type}}  & \multicolumn{6}{c}{Supervision Strategies}    \\  \cline{2-7}      & InfoNCE & FD & SD & KL-Div & Sym-SD & Sym-KL-Div             \\ \midrule \midrule
  intra-modal stu-stu learning   & $\backslash$ & $\backslash$  & \makecell[c]{$\mathcal{L}^{\mathrm{SD}}_{F^{T}_{\mathrm{stu}}\rightarrow F^{T}_{\mathrm{stu}}\Leftrightarrow	 F^{T}_{\mathrm{tch}}\rightarrow F^{T}_{\mathrm{tch}}}$\\+$\mathcal{L}^{\mathrm{SD}}_{F^{I}_{\mathrm{stu}}\rightarrow F^{I}_{\mathrm{stu}}\Leftrightarrow	 F^{I}_{\mathrm{tch}}\rightarrow F^{I}_{\mathrm{tch}}}$} & \makecell[c]{$\mathcal{L}^{\mathrm{KL\text{-}Div}}_{F^{T}_{\mathrm{stu}}\rightarrow F^{T}_{\mathrm{stu}}\lVert F^{T}_{\mathrm{tch}}\rightarrow F^{T}_{\mathrm{tch}}}$\\+$\mathcal{L}^{\mathrm{KL\text{-}Div}}_{F^{I}_{\mathrm{stu}}\rightarrow F^{I}_{\mathrm{stu}}\lVert F^{I}_{\mathrm{tch}}\rightarrow F^{I}_{\mathrm{tch}}}$} & $\mathcal{L}^{\mathrm{SD}}_{F^{T}_{\mathrm{stu}}\rightarrow F^{T}_{\mathrm{stu}}\Leftrightarrow	 F^{I}_{\mathrm{stu}}\rightarrow F^{I}_{\mathrm{stu}}}$ & \makecell[c]{$\mathcal{L}^{\mathrm{KL\text{-}Div}}_{F^{T}_{\mathrm{stu}}\rightarrow F^{T}_{\mathrm{stu}}\lVert F^{I}_{\mathrm{stu}}\rightarrow F^{I}_{\mathrm{stu}}}$\\+$\mathcal{L}^{\mathrm{KL\text{-}Div}}_{F^{I}_{\mathrm{stu}}\rightarrow F^{I}_{\mathrm{stu}}\lVert F^{T}_{\mathrm{stu}}\rightarrow F^{T}_{\mathrm{stu}}}$} \\ \midrule   
  inter-modal stu-stu learning   & \makecell[c]{$\mathcal{L}^{\mathrm{InfoNCE}}_{F^{T}_{\mathrm{stu}}\rightarrow F^{I}_{\mathrm{stu}}}$\\+$\mathcal{L}^{\mathrm{InfoNCE}}_{F^{I}_{\mathrm{stu}}\rightarrow F^{T}_{\mathrm{stu}}}$}  & $\mathcal{L}^{\mathrm{FD}}_{F^{T}_{\mathrm{stu}}\Leftrightarrow F^{I}_{\mathrm{stu}}}$ & \makecell[c]{$\mathcal{L}^{\mathrm{SD}}_{F^{T}_{\mathrm{stu}}\rightarrow F^{I}_{\mathrm{stu}}\Leftrightarrow	 F^{T}_{\mathrm{tch}}\rightarrow F^{I}_{\mathrm{tch}}}$\\+$\mathcal{L}^{\mathrm{SD}}_{F^{I}_{\mathrm{stu}}\rightarrow F^{T}_{\mathrm{stu}}\Leftrightarrow	 F^{I}_{\mathrm{tch}}\rightarrow F^{T}_{\mathrm{tch}}}$} &  \makecell[c]{$\mathcal{L}^{\mathrm{KL\text{-}Div}}_{F^{T}_{\mathrm{stu}}\rightarrow F^{I}_{\mathrm{stu}}\lVert F^{T}_{\mathrm{tch}}\rightarrow F^{I}_{\mathrm{tch}}}$\\+$\mathcal{L}^{\mathrm{KL\text{-}Div}}_{F^{I}_{\mathrm{stu}}\rightarrow F^{T}_{\mathrm{stu}}\lVert F^{I}_{\mathrm{tch}}\rightarrow F^{T}_{\mathrm{tch}}}$} & $\backslash$  & $\backslash$  \\    \midrule  
  intra-modal tch-stu learning   & \makecell[c]{$\mathcal{L}^{\mathrm{InfoNCE}}_{F^{T}_{\mathrm{stu}}\rightarrow F^{T}_{\mathrm{tch}}}$\\+$\mathcal{L}^{\mathrm{InfoNCE}}_{F^{I}_{\mathrm{stu}}\rightarrow F^{I}_{\mathrm{tch}}}$} & \makecell[c]{$\mathcal{L}^{\mathrm{FD}}_{F^{T}_{\mathrm{stu}}\Leftrightarrow F^{T}_{\mathrm{tch}}}$\\+$\mathcal{L}^{\mathrm{FD}}_{F^{I}_{\mathrm{stu}}\Leftrightarrow F^{I}_{\mathrm{tch}}}$} &  \makecell[c]{$\mathcal{L}^{\mathrm{SD}}_{F^{T}_{\mathrm{stu}}\rightarrow F^{T}_{\mathrm{tch}}\Leftrightarrow	 F^{T}_{\mathrm{tch}}\rightarrow F^{T}_{\mathrm{tch}}}$\\+$\mathcal{L}^{\mathrm{SD}}_{F^{I}_{\mathrm{stu}}\rightarrow F^{I}_{\mathrm{tch}}\Leftrightarrow	 F^{I}_{\mathrm{tch}}\rightarrow F^{I}_{\mathrm{tch}}}$} &  \makecell[c]{$\mathcal{L}^{\mathrm{KL\text{-}Div}}_{F^{T}_{\mathrm{stu}}\rightarrow F^{T}_{\mathrm{tch}}\lVert F^{T}_{\mathrm{tch}}\rightarrow F^{T}_{\mathrm{tch}}}$\\+$\mathcal{L}^{\mathrm{KL\text{-}Div}}_{F^{I}_{\mathrm{stu}}\rightarrow F^{I}_{\mathrm{tch}}\lVert F^{I}_{\mathrm{tch}}\rightarrow F^{I}_{\mathrm{tch}}}$} & $\mathcal{L}^{\mathrm{SD}}_{F^{T}_{\mathrm{stu}}\rightarrow F^{T}_{\mathrm{tch}}\Leftrightarrow	 F^{I}_{\mathrm{stu}}\rightarrow F^{I}_{\mathrm{tch}}}$  & \makecell[c]{$\mathcal{L}^{\mathrm{KL\text{-}Div}}_{F^{T}_{\mathrm{stu}}\rightarrow F^{T}_{\mathrm{tch}}\lVert F^{I}_{\mathrm{stu}}\rightarrow F^{I}_{\mathrm{tch}}}$\\+$\mathcal{L}^{\mathrm{KL\text{-}Div}}_{F^{I}_{\mathrm{stu}}\rightarrow F^{I}_{\mathrm{tch}}\lVert F^{T}_{\mathrm{stu}}\rightarrow F^{T}_{\mathrm{tch}}}$} \\   \midrule   
  inter-modal tch-stu learning   & \makecell[c]{$\mathcal{L}^{\mathrm{InfoNCE}}_{F^{T}_{\mathrm{stu}}\rightarrow F^{I}_{\mathrm{tch}}}$\\+$\mathcal{L}^{\mathrm{InfoNCE}}_{F^{I}_{\mathrm{stu}}\rightarrow F^{T}_{\mathrm{tch}}}$} & \makecell[c]{$\mathcal{L}^{\mathrm{FD}}_{F^{T}_{\mathrm{stu}}\Leftrightarrow F^{I}_{\mathrm{tch}}}$\\+$\mathcal{L}^{\mathrm{FD}}_{F^{I}_{\mathrm{stu}}\Leftrightarrow F^{T}_{\mathrm{tch}}}$} & \makecell[c]{$\mathcal{L}^{\mathrm{SD}}_{F^{T}_{\mathrm{stu}}\rightarrow F^{I}_{\mathrm{tch}}\Leftrightarrow	 F^{T}_{\mathrm{tch}}\rightarrow F^{I}_{\mathrm{tch}}}$\\+$\mathcal{L}^{\mathrm{SD}}_{F^{I}_{\mathrm{stu}}\rightarrow F^{T}_{\mathrm{tch}}\Leftrightarrow	 F^{I}_{\mathrm{tch}}\rightarrow F^{T}_{\mathrm{tch}}}$} &  \makecell[c]{$\mathcal{L}^{\mathrm{KL\text{-}Div}}_{F^{T}_{\mathrm{stu}}\rightarrow F^{I}_{\mathrm{tch}}\lVert F^{T}_{\mathrm{tch}}\rightarrow F^{I}_{\mathrm{tch}}}$\\+$\mathcal{L}^{\mathrm{KL\text{-}Div}}_{F^{I}_{\mathrm{stu}}\rightarrow F^{T}_{\mathrm{tch}}\lVert F^{I}_{\mathrm{tch}}\rightarrow F^{T}_{\mathrm{tch}}}$} & $\mathcal{L}^{\mathrm{SD}}_{F^{T}_{\mathrm{stu}}\rightarrow F^{I}_{\mathrm{tch}}\Leftrightarrow	 F^{I}_{\mathrm{stu}}\rightarrow F^{T}_{\mathrm{tch}}}$ & \makecell[c]{$\mathcal{L}^{\mathrm{KL\text{-}Div}}_{F^{T}_{\mathrm{stu}}\rightarrow F^{I}_{\mathrm{tch}}\lVert F^{I}_{\mathrm{stu}}\rightarrow F^{T}_{\mathrm{tch}}}$\\+$\mathcal{L}^{\mathrm{KL\text{-}Div}}_{F^{I}_{\mathrm{stu}}\rightarrow F^{T}_{\mathrm{tch}}\lVert F^{T}_{\mathrm{stu}}\rightarrow F^{I}_{\mathrm{tch}}}$}  \\    
\bottomrule 
\end{tabular}}
\caption{Detailed loss functions of all combinations of knowledge interaction learning and supervision strategies. "Sym-" is the symmetric version loss function. 
 "$\backslash$" indicates the combination is meaningless. 
}
\label{tab:lossfunc}
\end{table*}

It is worth noting that, when performing the learning process indicated by an arrow shown in Fig.~\ref{pipeline}, 
the common practice is to use teachers' outputs  $F^{T}_{\mathrm{tch}}$ and $F^{I}_{\mathrm{tch}}$ as target in Eq.~(\ref{sd})(\ref{kldiv}) that students learn from. 
While in our case with two modalities available,  we propose to use the paired arrow as the target, and we call this the \textbf{symmetric version} (for SD loss and KL-Div loss). For example, 
inter-modal teacher-student learning implemented with  KL-Div loss can be formulated as 
\begin{align}
\mathcal{L}^{\mathrm{KL\text{-}Div}}_{F^{T}_{\mathrm{stu}}\rightarrow F^{I}_{\mathrm{tch}}\lVert F^{T}_{\mathrm{tch}}\rightarrow F^{I}_{\mathrm{tch}}} + \mathcal{L}^{\mathrm{KL\text{-}Div}}_{F^{I}_{\mathrm{stu}}\rightarrow F^{T}_{\mathrm{tch}}\lVert F^{I}_{\mathrm{tch}}\rightarrow F^{T}_{\mathrm{tch}}},
\end{align}
while its symmetric version is 
\begin{align}
\mathcal{L}^{\mathrm{KL\text{-}Div}}_{F^{T}_{\mathrm{stu}}\rightarrow F^{I}_{\mathrm{tch}}\lVert F^{I}_{\mathrm{stu}}\rightarrow F^{T}_{\mathrm{tch}}} + \mathcal{L}^{\mathrm{KL\text{-}Div}}_{F^{I}_{\mathrm{stu}}\rightarrow F^{T}_{\mathrm{tch}}\lVert F^{T}_{\mathrm{stu}}\rightarrow F^{I}_{\mathrm{tch}}}.
\end{align}
This modification deepens the interaction between the four encoders during optimization. 

So far, 
any one of the learning types can be concretely 
 implemented by any one of the supervision strategies,
 except for a few  meaningless combinations. Detailed loss functions are listed in Tab.~\ref{tab:lossfunc}.

\section{Experiments}
\subsection{Setup}\label{setup}
We use Conceptual Caption (CC3M) \cite{cc3m} and Conceptual 12M  (CC12M) \cite{changpinyo2021cc12m} for pre-training distillation, which consist of 3M and 12M noisy text-image  pairs respectively.  
During fine-tuning, we use  MSCOCO \cite{coco} and Flickr30K \cite{plummer2015flickr30k} as benchmarks. 
MSCOCO has 113,287 images for training, 5K images for validation, and both 5K and 1K for testing. Flickr30K has 28,783 images for training, 1K images for validation, and 1K for testing.
Following previous works, we use recall R@\textit{k} (\textit{k}=1,5,10) as the main metric.

We use the open-source CLIP \cite{clip} with ViT-B/32 \cite{vit} as the teacher model.
Its image encoder is a 12-layer ViT with the hidden size to be 768 and 12 attention heads. Its text encoder is a 12-layer Transformer with  hidden size to be 512 and 8 attention heads. 

For the student model, we use ViT-S/16 with hidden size to be 384 as the image encoder, 
 and  initialize it from  the pre-trained weights on ImageNet-21K \cite{ridnik2021imagenet21k}. 
 For the text encoder, we experiment with 2, 4 and 6-layer Transformer, of which the weights are initialized from the first corresponding layers of the teacher's text encoder.
The details of model settings are shown in Tab.~\ref{tab:modelset}.

In pre-training distillation, we train the student models in 4 epochs using AdamW \cite{adamw} with a batch size of 1024 for both images and texts, the learning rate of 3e-4, and the weight decay of 0.1.
We employ a cosine learning rate scheduler with 10,000 warm-up steps. 
In fine-tuning, we  use the same optimization setting as in MoTIS \cite{motis}.
Experiments are conducted on 4 NVIDIA TESLA V100 32G GPUs.

\begin{table*}[!]
\centering
\scalebox{0.77}{
\begin{tabular}{@{}l|ccccccccccc@{}}
\toprule
 \multicolumn{1}{c|}{\multirow{2}{*}{Learning Type}}  & \multicolumn{6}{c}{Supervision Strategies}    \\  \cline{2-7}      & InfoNCE & FD & SD & KL-Div & Sym-SD & Sym-KL-Div             \\ \midrule \midrule
  intra-modal stu-stu learning   & $\backslash$ & $\backslash$  & \textbf{58.8/83.7/90.1} & 57.1/82.7/88.8 & 57.1/82.0/89.2 & 56.8/81.6/88.6 \\    \midrule
  inter-modal stu-stu learning   & 34.7/58.7/69.9 & 56.6/82.1/88.8 & \textbf{58.6/83.6/90.0} &  56.5/82.4/88.9& $\backslash$  & $\backslash$  \\    \midrule
  intra-modal tch-stu learning   & 57.6/82.4/89.0$^{\dag}$ &57.6/82.0/88.4 &  \textbf{58.5/83.2/89.6} & 55.1/80.0/87.4&  \textbf{58.7/83.4/89.9} & 56.3/81.5/88.3 \\    \midrule
  inter-modal tch-stu learning   & 51.4/76.3/83.8 &50.0/80.7/88.4 & 57.6/82.5/88.6 & 56.9/81.8/88.7 & 56.9/81.8/88.7 &  \textbf{59.1/83.4/89.8} \\    
\bottomrule 
\end{tabular}}
\caption{
Ablation study  of text-image retrieval R@1/5/10 on Flickr30K. $^{\dag}$Baseline. \textbf{Bold} denotes all R@\textit{k}s have obvious improvements. All five losses \textbf{in bold} will be added to the baseline loss to finally serve as our framework.
}
\label{tab:ablation}
\end{table*}

\begin{table*}[!]
\centering
\scalebox{0.75}{
\begin{tabular}{@{}l|c|c|ccccccccc@{}}
\toprule
 \multicolumn{1}{c|}{\multirow{2}{*}{Model}} &  \multirow{2}{*}{\makecell[c]{Text\\Encoder}} &  \multirow{2}{*}{\makecell[c]{Image\\Encoder}} & \multicolumn{3}{c}{Flickr30K} & \multicolumn{3}{c}{MSCOCO (1K)} & \multicolumn{3}{c}{MSCOCO (5K)}    \\ \cmidrule(l){4-6}  \cmidrule(l){7-9} \cmidrule(l){10-12}  &    &  &R@1&R@5&R@10&  R@1&R@5&R@10& R@1&R@5&R@10   \\ \midrule \midrule
 \multicolumn{12}{l}{(a) \textit{Fair Comparisons}}\\
 \midrule
InfoNCE-based &\multicolumn{1}{c|}{\multirow{4}{*}{CLIP's[512/6]}}  &\multicolumn{1}{c|}{\multirow{4}{*}{ViT-S/16[384/12]}} &38.4	& 68.0	& 78.0 & 53.3	& 85.3	& 93.5 & 31.5	&60.3	& 73.3 \\
Cross-modal KD &  & &41.1	&70.6& 80.0 &54.9	& 86.0	& 93.6 & 33.4	&61.9 & 74.4 \\
MoTIS & & &57.0 & 82.1&88.8 &62.7&88.2 &94.5&42.6 &69.6& 79.4\\
ConaCLIP \textbf{(Ours)} & & & \textbf{60.6}  &\textbf{85.2} & \textbf{91.2} & \textbf{68.6} & \textbf{92.4} & \textbf{96.7} & \textbf{47.3}  & \textbf{76.1 }&  \textbf{85.2}  \\
\midrule
 \multicolumn{12}{l}{(b) \textit{Model Zoo and Benchmarks}}\\
\midrule
ConaCLIP$\texttt{-6L}$ \textbf{(Ours) }& CLIP's[512/6]&\multicolumn{1}{c|}{\multirow{3}{*}{ViT-S/16[384/12]}} &  \textbf{67.6} & \textbf{89.6} & \textbf{94.4} &  \textbf{75.6}   & \textbf{94.6}  & \textbf{97.4} & \textbf{55.4} &   \textbf{83.5} & \textbf{89.9} \\
ConaCLIP$\texttt{-4L}$ \textbf{(Ours) }&CLIP's[512/4] & & 67.0  & 89.3  & 94.2 &75.4 & \textbf{94.6}  & \textbf{97.4} & 55.3   &83.1 & \textbf{89.9} \\
ConaCLIP$\texttt{-2L}$ \textbf{(Ours)} &CLIP's[512/2] & &65.6   & 89.2 & 93.9 & 74.7  & 94.3 & 97.3 & 54.1    & 82.2  & 89.4 \\
\bottomrule 
\end{tabular}}
\caption{(a) Fair comparisons of text-image retrieval results on Flickr30K and  MSCOCO (1K and 5K).  
(b) Our model zoo and the corresponding benchmarks.
\textbf{Bold} indicates the best performance. 
"[\textit{m}/\textit{n}]" represents \textit{n} layers with the hidden size to be \textit{m}.
}
\label{tab:sota}
\end{table*}

\subsection{Ablation Study}\label{ablation}
Considering our complete pre-training distillation takes a relatively long
time, we follow the setup of \cite{motis} and train ConaCLIP on CC3M for 1 epoch with batch size 84 to conduct the ablation study. 
Taking Eq.~(\ref{equal_motis}) as the naive baseline, we aim to find out which of the proposed combinations of learning types and supervision strategies can bring further  improvements.
The fine-tuned results on Flickr30K is shown in Tab.~\ref{tab:ablation}.

We can make some observations that: 
1) With an appropriate choice of detailed supervision strategies, each type of learning can further bring obvious improvements on the basis of the baseline. 
2) The effect of each learning type is greatly affected by the implemented loss function. It also indicates that the pre-training distillation process should be carefully explored regarding the supervision strategy. 
3) Our proposed symmetric version losses (Sym-SD and Sym-KL-Div)  can generally achieve superior performances to the standard ones for (intra/inter-modal) teacher-student learning.

We can also attain several findings that:
1) For (intra/inter-modal) student-student learning where students first make knowledge interaction and then learn together from teachers, SD loss performs the best. 
Because the actual retrieval application uses this cosine similarity to rank candidates, it can help students acquire  goal-oriented knowledge more directly. It also relaxes the learning task of  students  from teachers' feature space to the similarity space.
2) For (intra/inter-modal) teacher-student learning, our proposed symmetric version losses are more suitable. 
Compared with the standard losses, they make the knowledge interaction between teachers and students  closer during optimization. In this regard, student encoders can cooperate more intimately in downstream tasks.
3) Although the naive intra-modal teacher-student learning with InfoNCE loss can  serve as a competent baseline, 
the addition of SD and Sym-SD losses of the same learning type can  complement its effectiveness.
On the other hand, 
the other three different  learning  types with  proper loss choices can also benefit the effect of pre-training distillation.
More findings on distilling intermediate layers are shown in \ref{MiddleLayer}.

Our method has been established with the further integration of the highlight (in bold) combinations in Tab.~\ref{tab:ablation} based on the baseline. 
The  effect after full integration is shown in Tab.~\ref{tab:sota}(a).

\begin{table*}[!]
\centering
\scalebox{0.92}{
\begin{tabular}{@{}l|ccc|ccc|ccccc@{}}
\toprule
 \multicolumn{1}{c|}{\multirow{2}{*}{Model}}  & \multicolumn{3}{c|}{Text-Image Retrieval} & \multicolumn{3}{c|}{Image-Text Retrieval}  & \multirow{2}{*}{Disk Space (MB)}  & \multirow{2}{*}{ QPS$_t$ }&  \multirow{2}{*}{QPS$_i$}\\ \cmidrule(l){2-4} \cmidrule(l){5-7}
  & R@1& R@5& R@10& R@1& R@5& R@10 & & & \\ \midrule \midrule
CLIP  &16.5 &48.0 & 61.3 & 18.0& 49.7  & 62.2 & 578 &  1.00$\times$  & 1.00$\times$ \\
EC-CLIP  &\textbf{25.0} & \textbf{63.5}& \textbf{76.0}  &\textbf{25.9}& \textbf{64.1} & \textbf{75.7}    & 578 &  1.00$\times$  &  1.00$\times$ \\ \midrule
EC-ConaCLIP$\texttt{-6L}$  &24.3&62.4&75.6  &24.8&62.4&73.9    & 254   &  1.92$\times$  &  \textbf{1.44$\times$}\\
EC-ConaCLIP$\texttt{-4L}$  &23.6&61.1&74.3  &22.0&59.7&72.2    & 230   &  2.71$\times$ &  \textbf{1.44$\times$}\\
EC-ConaCLIP$\texttt{-2L}$  &23.0&60.7&73.5  &21.8&59.3&72.0    & \textbf{206}  &  \textbf{4.86$\times$} &  \textbf{1.44$\times$} \\
\bottomrule 
\end{tabular}}
\caption{
Performance of the industry application.   
"EC-" is the e-commercial version of our model. 
QPS$_t$/QPS$_i$ indicates the acceleration rate of QPS.
}
\label{tab:industry}
\end{table*}

\subsection{Performance}
\paragraph{Fair Comparisons.} In order to better verify the effectiveness of ConaCLIP, besides the previous SOTA, we also experiment with two strong baseline methods.
As shown in Tab.~\ref{tab:sota}(a), \textit{InfoNCE-based} indicates the naive cross-modal contrastive learning procedure. 
\textit{Cross-modal KD} represents 
distilling the cross-modal in-batch probability distribution  of teachers into students.
All these experiments are conducted under the pre-training setup of \cite{motis} for fair comparisons.
As can be observed, 
1) Cross-modal KD which introduces the knowledge distillation process obviously outperforms the standard InfoNCE-based approach.
2) MoTIS greatly surpasses InfoNCE-based and Cross-modal KD. This reveals the superiority of intra-modal teacher-student learning over inter-modal student-student learning in the case of dual-encoder distillation.
3) Our ConaCLIP shows significant improvements  compared with competitors on all evaluation metrics: 3.6/3.1/2.4 R@1/5/10 gains on Flickr30K, 5.9/4.2/2.2 R@1/5/10 gains on MSCOCO (1K) and 4.7/6.5/5.8 R@1/5/10 gains on MSCOCO (5K).  This fully demonstrates the effectiveness of our distillation framework with Cona.

\paragraph{Model Zoo and Benchmarks.} 
In order to better promote the development of cross-modal text-image research, we release a series of lightweight dual-encoder models. Their benchmark results are shown in Tab.~\ref{tab:sota}(b).
In this case, the power of ConaCLIP is further unlocked and brings further improvements. 
Specifically, even ConaCLIP$\texttt{-2L}$ can achieve 
8.6/7.1/5.1 R@1/5/10 gains on Flickr30K, 12.0/6.1/2.8 R@1/5/10 gains on MSCOCO (1K) and 11.5/12.6/10.0 R@1/5/10 gains on MSCOCO (5K)  
compared with the previous SOTA.
We have also found that the capacity of the text encoder may  have limited effects on these performances. 
For example, ConaCLIP$\texttt{-4L}$ can achieve competitive results with ConaCLIP$\texttt{-6L}$, and ConaCLIP$\texttt{-2L}$ has only minor drops.

\section{Industry Application}
We apply the proposed technique to end-to-end cross-modal retrieval in an e-commerce platform, where we vectorize the search queries and the products and then perform  product retrieval and ranking with nearest-neighbor search \cite{muja2009fast,jegou2010product,johnson2019billion}, as shown in Fig.~\ref{app}. 
We first collect massive data of text-image pairs from e-commerce  products in our platform, 
 where the titles of products can act as text  information. 
 % In total, 
 We utilize most of the data to pre-train an e-commerce version of the CLIP model (denoted as EC-CLIP) with ViT-B/32 as the image encoder, which is overly large for online deployment.
For the remaining data, we  utilize 3M pairs for distilling the lightweight EC-ConaCLIP.
To evaluate its effectiveness, we hold out a separate set of 100K pairs for fine-tuning and 5K/5K pairs used in validating/testing.
In this set of experiments, 
we train EC-ConaCLIP  for 20 epochs in pre-training distillation, and fine-tune both EC-CLIP and EC-ConaCLIP for 5 epochs. The remaining settings are  the same as in  Section \ref{setup}.

In apart to the R@\textit{k} metric, we also report the disk space (MB) and the acceleration rate of Query Per Second
(QPS$_i$ for image and QPS$_t$ for text) 
to evaluate model’s memory footprints and inference speed.
In Tab.~\ref{tab:industry}, we report the averaged results  where the inference speed is tested on an NVIDIA TESLA V100 (16G) GPU.
As seen, the compressed EC-ConaCLIP$\texttt{-6L}$  only takes 44\% disk space (254MB) of EC-CLIP meanwhile being 1.44$\times$/1.92$\times$ faster with image/text queries.
It also performs on par with EC-CLIP.
Our EC-ConaCLIP$\texttt{-2L}$  can further achieve up to 4.86$\times$ inference speed-up with text queries, and 64\% size reduction (from 578MB to 206MB).
 We provide some case studies in \ref{CaseStudy}.

\section{Conclusion}
In this paper, we propose 
Cona
for pre-training distillation with dual-encoder architecture.
It gathers every type of knowledge interaction learning with appropriate  supervision choice to benefit the cross-modal distillation. 
The resulting ConaCLIP achieves superior performances on both general benchmarks and industry applications.  

For future work, we will explore more variants of visual encoders, and continue to tap the potential of dual-encoder distillation.

\section*{Acknowledgements}
This research is supported in part by Alibaba Innovative Research Foundation (No. D8200510), NSFC (Grant No. 61936003), Zhuhai Industry Core and Key Technology Research Project (No. 2220004002350).
This work is also supported by Alibaba Cloud Group,  through Research Talent Program with South China University of Technology.

\bibliography{anthology}
\bibliographystyle{acl_natbib}

\newpage
\appendix

\section{Appendix}
\label{sec:appendix}

\begin{table}[bp]
\centering
\scalebox{0.95}{
\begin{tabular}{@{}c|ccccccccccc@{}}
\toprule
 Applied Parts  & R@1 & R@5 & R@10    \\ \midrule \midrule
  6th (Baseline)    & \textbf{60.6} &   \textbf{85.2}  &  \textbf{91.2} \\  
  5-6   & 59.5 & 84.2  & 90.7  \\    
  4-6   & 59.5 & 84.3 & 90.9 \\    
  3-6   & 57.9  & 83.5  & 90.2 \\ 
  2-6  & 58.6 & 84.1  &  90.9 \\   
  1-6  & 59.2 & 84.5  & 90.8  \\   
\bottomrule 
\end{tabular}}
\caption{
An exploratory study on distilling intermediate layers. The R@1/5/10 results on Flickr30K are listed. Each student/teacher encoder is evenly divided into six parts along the number of layers, and distillation is performed on the feature representations of each part.
}
\label{tab:middlelayer}
\end{table}

\subsection{Model Settings}
We give  detailed parameters on the settings of our ConaCLIP models in Tab.~\ref{tab:modelset}, such as the number of parameters, layers, heads, etc.

\begin{table}[H]
\centering
\scalebox{0.58}{
\begin{tabular}{@{}l|ccccccccccc@{}}
\toprule
 Model Setting  & ConaCLIP$\texttt{-6L}$ & ConaCLIP$\texttt{-4L}$ & ConaCLIP$\texttt{-2L}$    \\ \midrule \midrule
  Number of Parameters   & 66M &  60M  &  53M  \\   \midrule 
  Text Encoder Layers   & 6 & 4  & 2  \\    
  Text Encoder Heads   & 8 & 8 & 8 \\    
  Text Encoder Hidden Size   & 512 & 512  & 512 \\    
  Vocabulary Size  & 49408 & 49408   &  49408\\   
  Text Length   & 77 & 77  & 77  \\   \midrule
  Image Encoder Layers   & 12 & 12  & 12  \\    
  Image Encoder Heads   & 6 & 6 & 6  \\    
  Image Encoder Hidden Size   & 384  & 384  & 384 \\   
  Image Patch Size   & 16 & 16  & 16  \\   
  Image Size   & 224 & 224  & 224 \\   
\bottomrule 
\end{tabular}}
\caption{
Detailed parameters on the settings of  our ConaCLIP models.
}
\label{tab:modelset}
\end{table}

\subsection{Negative Results on Distilling Intermediate Layers}\label{MiddleLayer}
We also present an exploratory study on distilling the knowledge of intermediate layers from teacher encoders. We first evenly divide the encoder of each student/teacher  into six parts along the number of layers, and then perform our  distillation  technique on the feature representations of each part. The experiment results are shown in Tab.~\ref{tab:middlelayer}. 

We can observe that additional distillation with  features of the intermediate layers does not bring about positive improvement.
This inspires us that we should mainly focus on the representation matching ability of the output of the last layer for the cross-modal retrieval task.
Due to the difference of capabilities between models of different sizes, they can choose different paths to learn the goal-oriented features in the same task during distillation \cite{li2021reskd,kdgap3,kdgap2}. In our application, we suggest that it can be inappropriate to force small models to learn the same path as the large ones.

\subsection{Application in E-Commerce Product Retrieval}
\begin{figure}[H]
\centering
\includegraphics[width=0.48\textwidth]{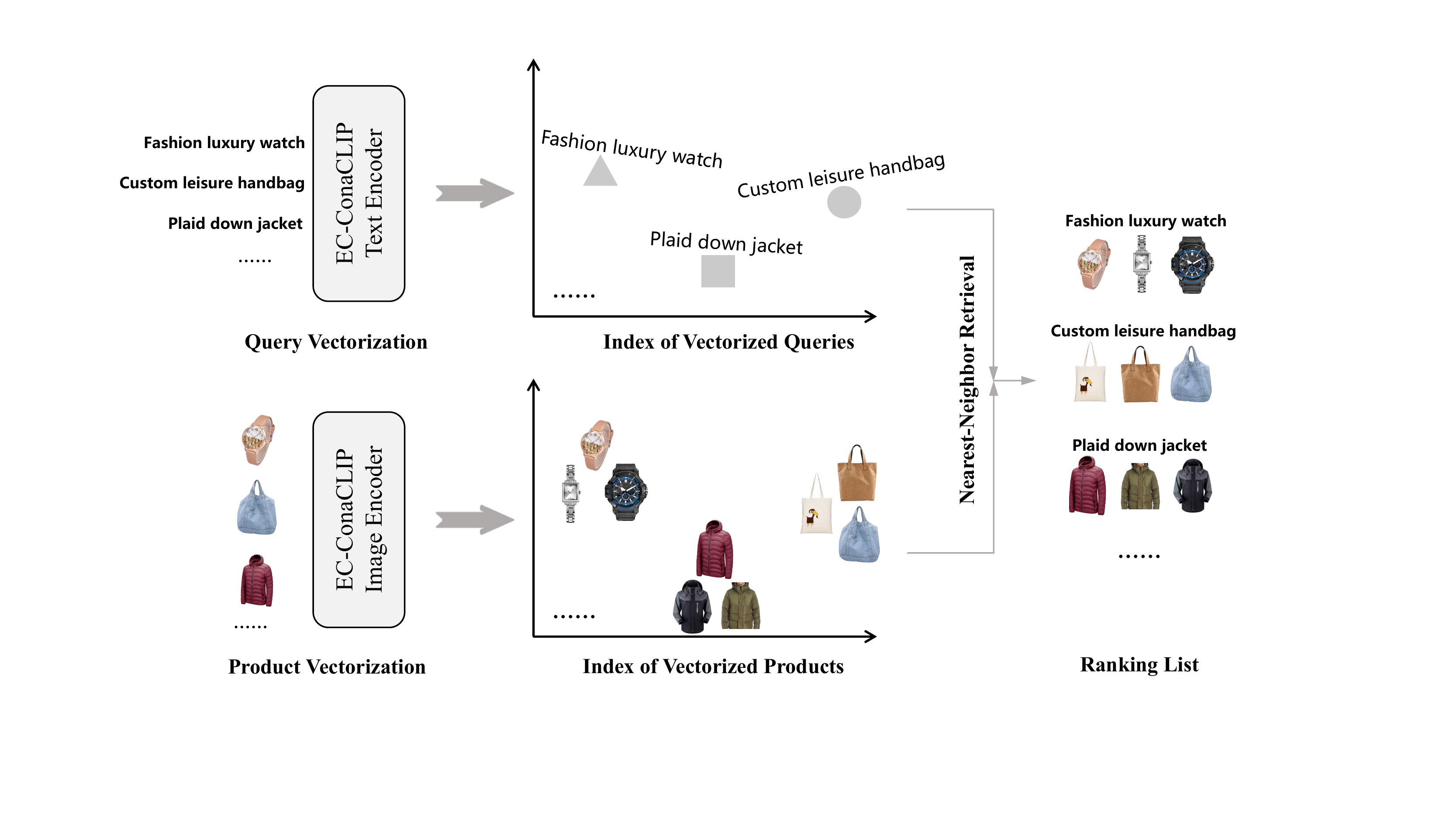}
\caption{The  application of our ConaCLIP in e-commerce retrieval.
} 
\label{app}
\end{figure}

We apply the proposed distillation technique to end-to-end cross-modal retrieval in an e-commerce platform, where we vectorize the search queries and the products and then perform  product retrieval and ranking with nearest-neighbor search. The whole framework is shown in Fig.~\ref{app}.

\subsection{Case Study}\label{CaseStudy}

\begin{table}[H]
\centering
\scalebox{0.62}{
\begin{tabular}{@{}cccccccccccc@{}}
\toprule
 Case  & Query & CLIP & EC-ConaCLIP \textbf{(Ours)}    \\ \midrule \midrule
  1   & \makecell[l]{Waterproof large capacity\\ lightweight fashion unicorn\\ cartoon kids girl middle \\school backpack.} &  \begin{minipage}[b]{0.3\columnwidth}\centering\raisebox{-.5\height}{\includegraphics[width=\linewidth]{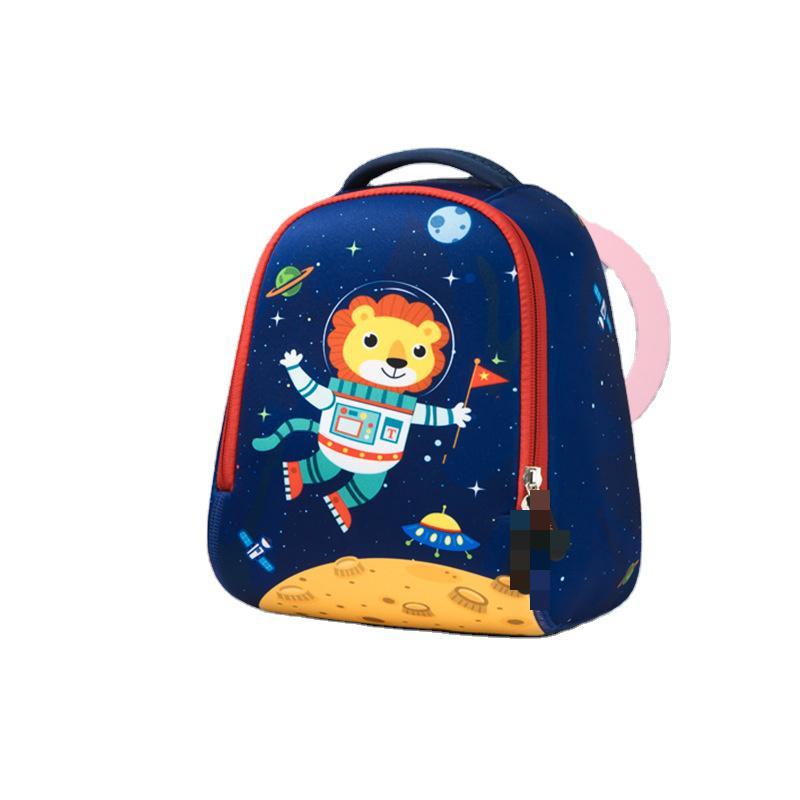}}\end{minipage}  &  \begin{minipage}[b]{0.3\columnwidth}\centering\raisebox{-.5\height}{\includegraphics[width=\linewidth]{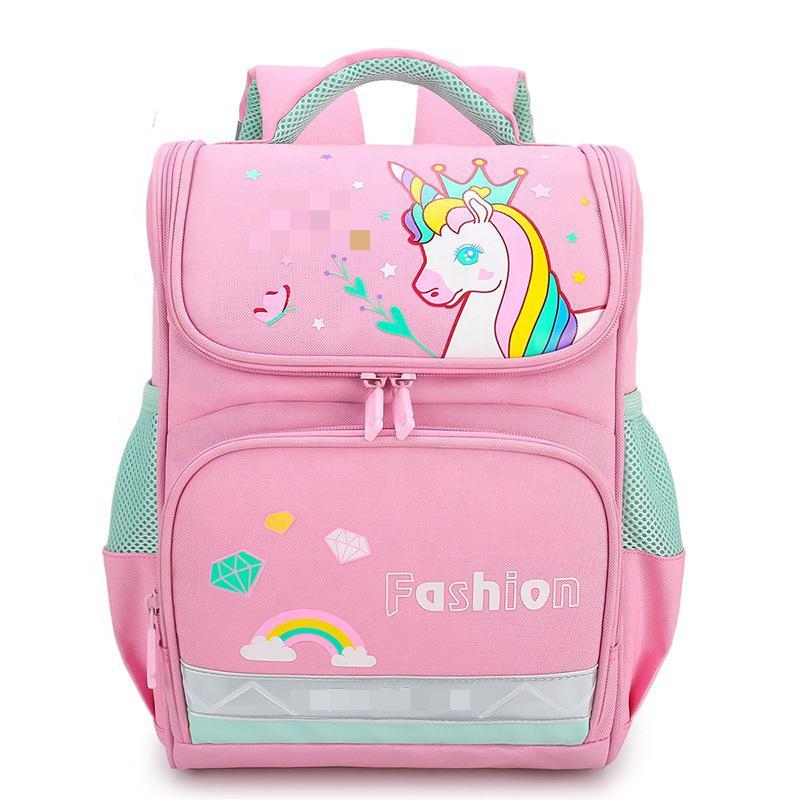}}\end{minipage}  \\ \midrule 
  2   & \makecell[l]{Stainless steel induction \\steamers pot, 2 layers\\ double handle food\\ cooking pots with lid.} & \begin{minipage}[b]{0.3\columnwidth}\centering\raisebox{-.5\height}{\includegraphics[width=\linewidth]{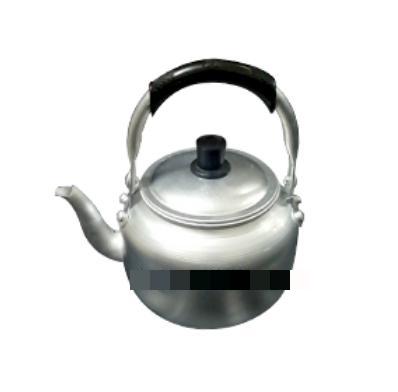}}\end{minipage}  & \begin{minipage}[b]{0.3\columnwidth}\centering\raisebox{-.5\height}{\includegraphics[width=\linewidth]{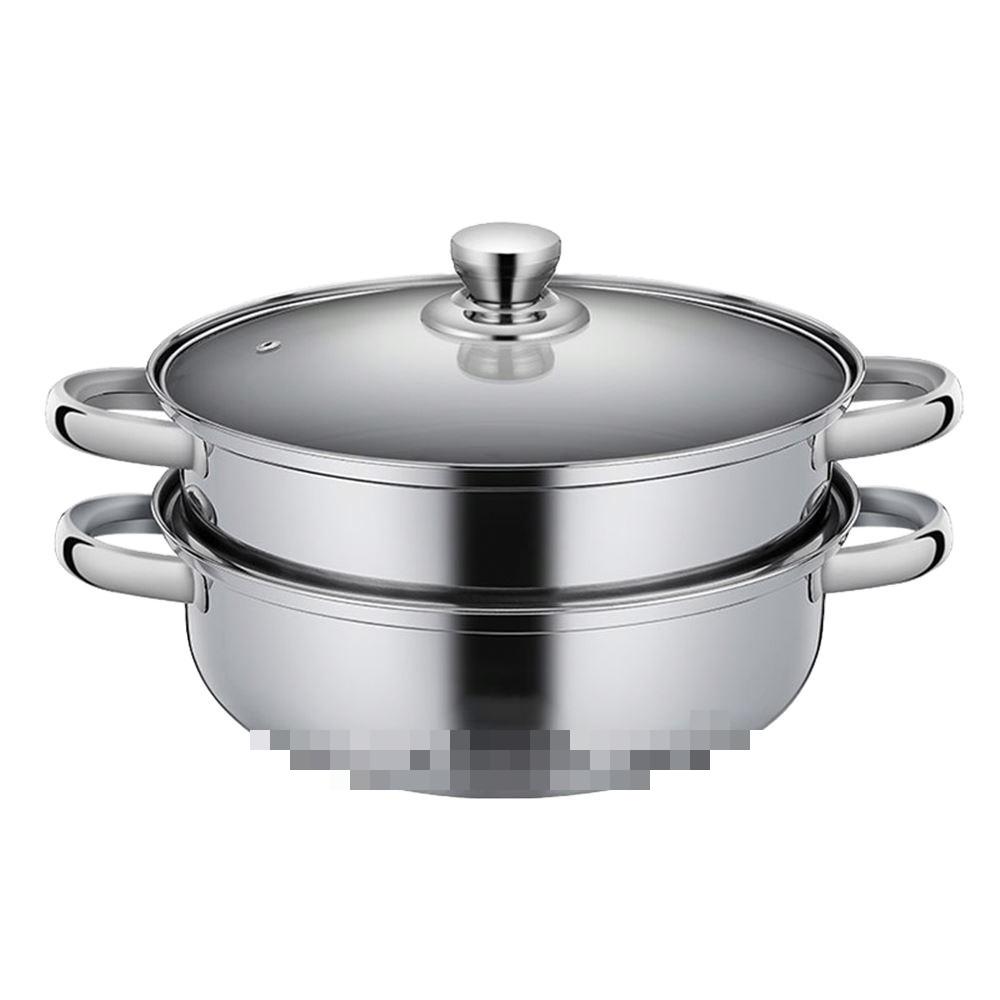}}\end{minipage}  \\    \midrule 
  3  & \makecell[l]{Children's sand hammer\\ wooden bell multi-color \\children's development toy.} & \begin{minipage}[b]{0.3\columnwidth}\centering\raisebox{-.5\height}{\includegraphics[width=\linewidth]{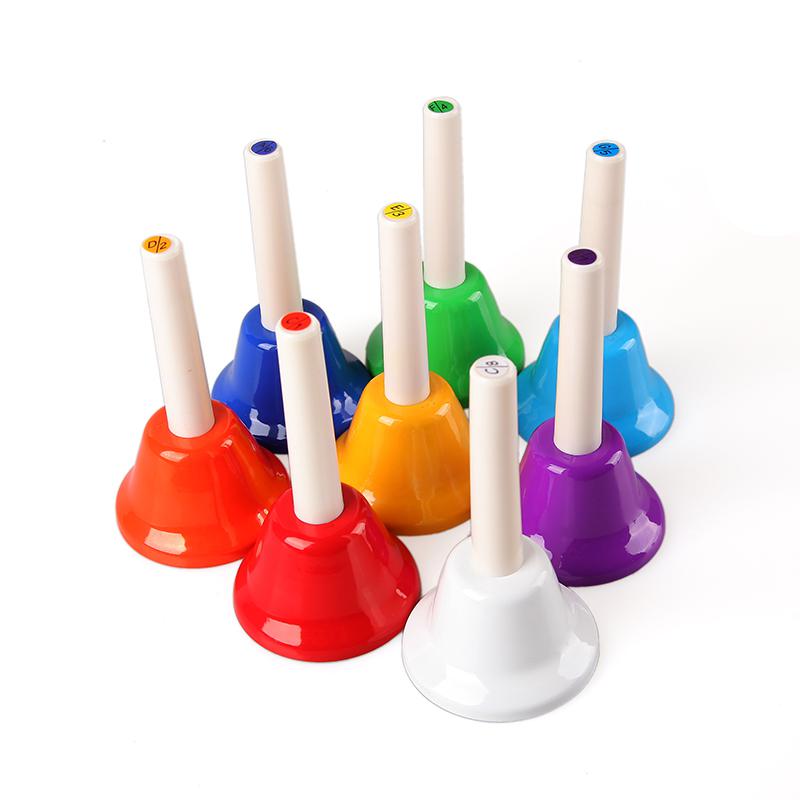}}\end{minipage} & \begin{minipage}[b]{0.3\columnwidth}\centering\raisebox{-.5\height}{\includegraphics[width=\linewidth]{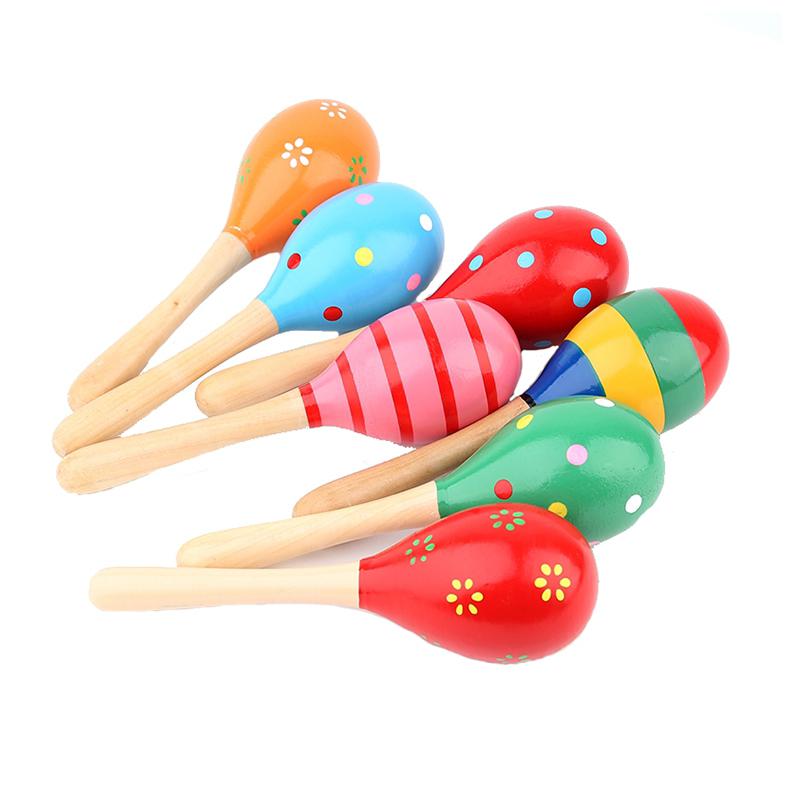}}\end{minipage} \\    \midrule 
  4   & \makecell[l]{Hot red large size sports\\ tights high waist yoga\\ pants.} & \begin{minipage}[b]{0.3\columnwidth}\centering\raisebox{-.5\height}{\includegraphics[width=\linewidth]{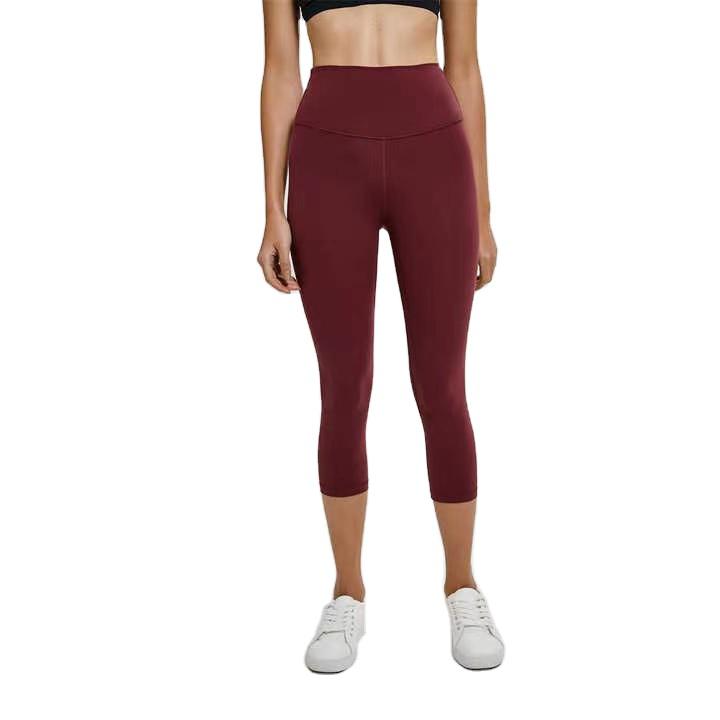}}\end{minipage}  & \begin{minipage}[b]{0.3\columnwidth}\centering\raisebox{-.5\height}{\includegraphics[width=\linewidth]{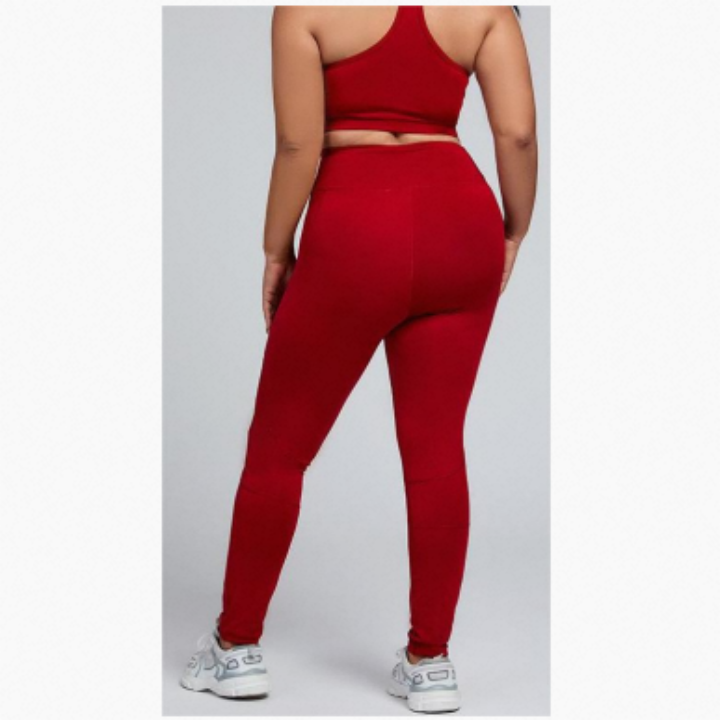}}\end{minipage} \\ \midrule 
  5 &  \makecell[l]{Tempered glass waterproof\\ platform 5kg digital food\\ electric kitchen scale.}  & \begin{minipage}[b]{0.3\columnwidth}\centering\raisebox{-.5\height}{\includegraphics[width=\linewidth]{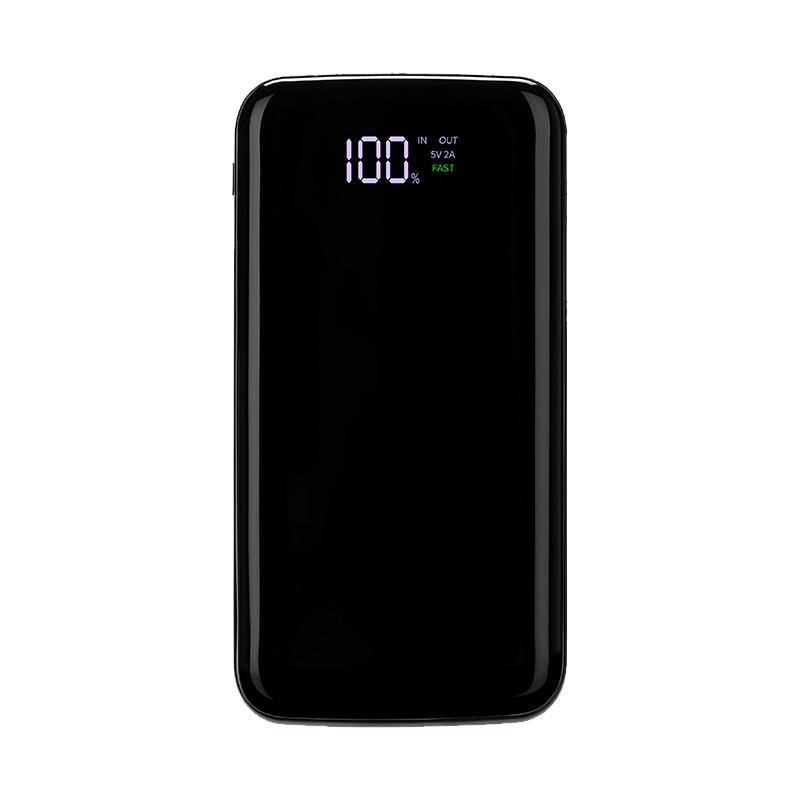}}\end{minipage}  &  \begin{minipage}[b]{0.3\columnwidth}\centering\raisebox{-.5\height}{\includegraphics[width=\linewidth]{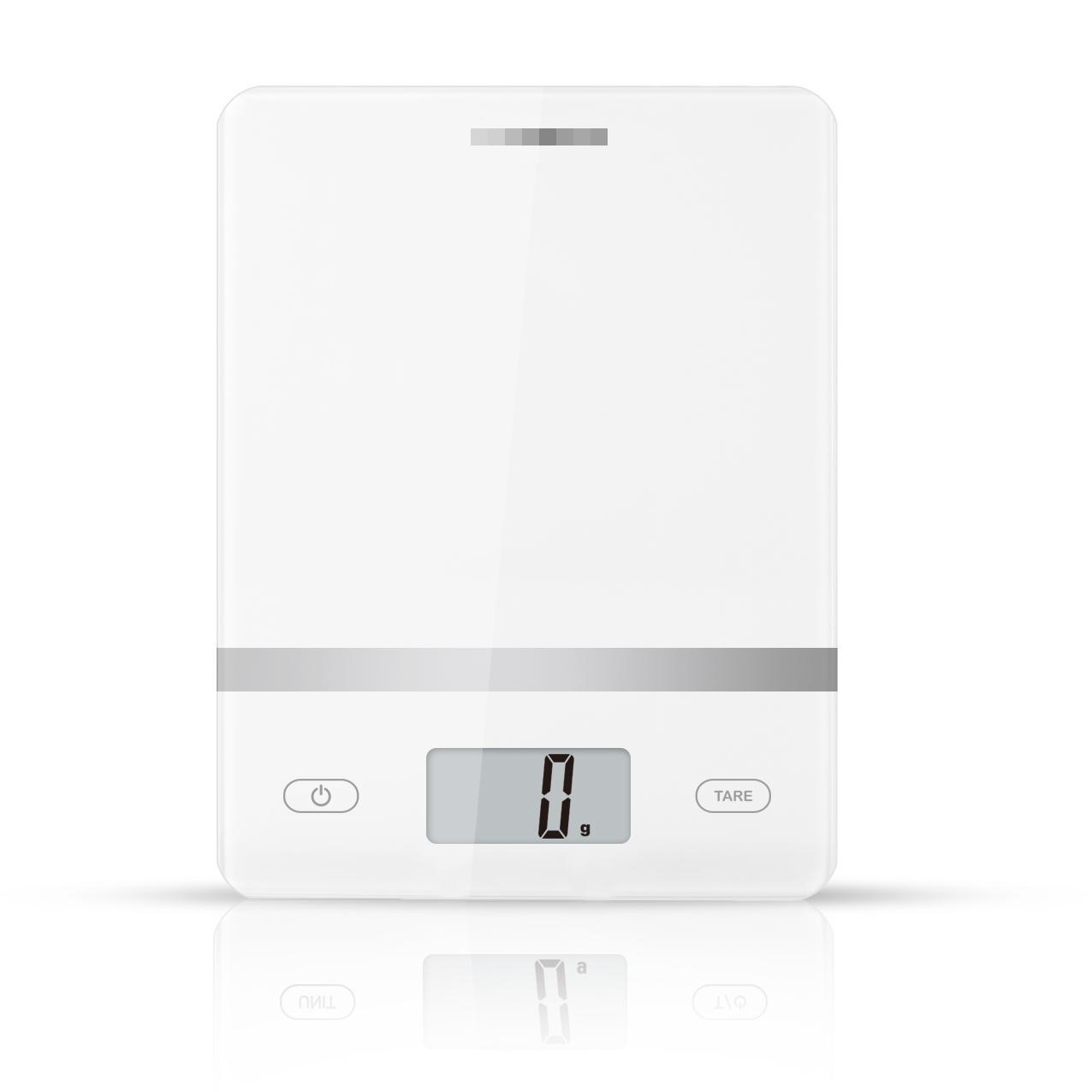}}\end{minipage} \\  
\bottomrule 
\end{tabular}}
\caption{
Case studies in e-commerce retrieval. Given the same text query, we show the image retrieval results of the open-source CLIP  and our EC-ConaCLIP.
}
\label{tab:casestudy}
\end{table}

Tab.~\ref{tab:casestudy} shows the case studies  in our e-commerce retrieval scenario. 
For the same text query, we show the top-1 image retrieval results of the open-source CLIP model and  our EC-ConaCLIP model respectively.

From these cases, we can find that our model can better capture conceptual and fine-grained fashion information during  cross-modal text-image retrieval, and  maintain the cross-modal alignment effect of text-image samples after the  lightweight distillation.
For example, in Case 1,  our model more accurately captures the cartoon subject in the target commodity as "unicorn".
In Case 2, our model pays more attention to fine-grained information "2 layers double handle", while maintaining the correct perception of other information such as "Stainless steel", "steamers pot" and "with lid". In Case 3, our EC-ConaCLIP better captures the color clue of "Hot red". Although the retrieval result of CLIP also conforms to the information of "sports tights high waist yoga pants", its color is more like "dark red".

Based on our distillation technique, the resulting model can sufficiently learn the perception ability of the teacher model about commodity fashion concepts and  reduce matching errors.

\end{document}